# BioMed-Agent-RL: A Meta Learning, All You Need for Biomedical Applications

Md Asaduzzaman Jabin, *Member, IEEE*, Zihao Wu, and Tianming Liu, *Member*, *IEEE*

***Abstract*—The current progress of Clinical Vision Large Language Models (C-VLLMs) has substantially improved digital diagnostics, still these frameworks often endure lesion noises, modality misalignment, hallucination, and missed contextual grounding in complex clinical cases. Moreover, prevailing agent systems usually depend on static and non-adaptable pipelines and lack the versatility necessary for complex medical reasoning. To resolve these difficulties, we present BioMed-Agent-RL, a unified medical agent that incorporates adaptive orchestration, policy, and reward-based reinforcement learning (RL) models for biomedical applications. To ensure reliability, it invokes clinical context-aware preference optimization (CPO), direct preference optimization (DPO), and group relative policy optimization (GRPO) with dynamic entropy regulation. This pipeline utilizes a multimodal meta-learning approach that operates as a field-specific expert and human judgment synthesizer. The agent adaptively utilizes a set of model-level expertise, such as clinical grounding and reasoner, lesion segmenter, and field-specific synthesizer, across various clinical modalities (e.g., X-ray) by utilizing an iterative and adaptive RL approach. The agent learns to seriously synthesize misleading, conflicting vision cues and trust in inherent reasoning, while specialist advice is faulty. An intensive ablation study is conducted across multiple benchmarks, and the agent significantly outperforms existing state of the art models, such as GPT-5, attaining up to ~73% accuracy (gain of ~5%) over contemporary baselines. As a result, the framework suggests a new standard for building factual, reliable, robust, and expert-like intelligent agent systems for independent clinical reasoning.**

## I. INTRODUCTION

THE recent progress of multimodal vision large language models (VLLMs) into biomedical applications [1] across various clinical modalities like MRI, CT, X-ray, and Gross, etc., has exhibited substantial possibilities for intelligent reasoning and clinical diagnostics. However, VLLMs trained on large multimodal clinical data depict promising performance to some extent, yet a single backbone often fails in reasoning [2] of various clinical subfields (e.g., radiology). Still, the advancement of clinical intelligence systems depicts several major challenges [3]- high computation cost associated with human-feedback alignment and optimization, the limited generalization capabilities of sole backbone across distinct modalities, hallucinations caused by multimodal misalignment, and non-adaptive, non-iterative essence of contemporary agent systems.

Achieving specific and most accurate iterative reasoning direction in foundation models generally entails reinforcement learning with human feedback (RLHF) with proximal policy optimization (PPO) [4], a method that is often computation-intensive and time consuming due to need of huge manual data annotation and tuning a huge distinct reward model. To tackle this, direct preference optimization (DPO) [5] has surfaced as firm and lightweight option which optimizes LLM models by drawing policy function in the loop through binary cross-entropy, bypass external reward model. While DPO elevates general contextual conversations, but expert-level clinical tasks necessitate further greater precision.

In the medical field, a generalist LLM backbone frequently struggles [6] to tackle the reasoning and lesion grounding task necessary for diverse branches like pathology, radiology, and surgery, etc. This requires an agentic framework, multimodal clinical agent, able to adaptively orchestrating specific mechanism [7] for multimodal grounding, lesion reasoning and segmentation, report and summary generation across various clinical cues like, MRI, CT, and X-rays. In spite of the triumph of such agents, they repeatedly endure from multimodal misalignment where the model mostly

Md Asaduzzaman Jabin, Zihao Wu, and Tianming Liu are with The University of Georgia, Athens, Georgia, GA 30605 USA. (e-mail: {mj71006, zihao.wu1, tliu}@uga.edu).

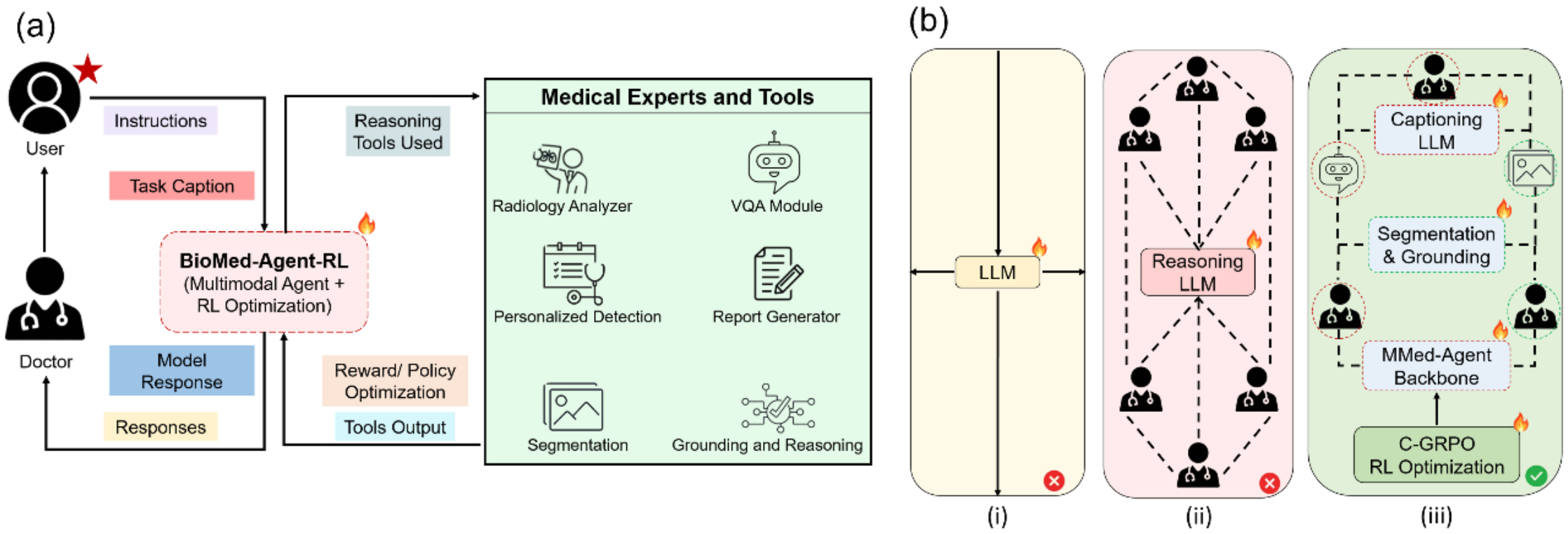

Fig. 1. The overview of BioMed-Agent-RL- (a) The reasoning loop; and (b) The agent comparison paradigm- (i) Single backbone agent, (ii) Multisystem agent, and (iii) C-GRPO BioMed-Agent-RL (Ours).

prefers text-based knowledge over visual embedding. Therefore, it leads to erroneous reasoning and fatal hallucinations. To alleviate these issues, clinical context preference optimization (CPO) [8] selects excellent-quality preference cues by presenting "rejected" instances through region-specific lesion noising. However, the noising often disturbs visual reasoning of vital investigative areas and causes the inclusion of hallucinations [9]. However, by applying weights over these cues through agent pipelines, the model grasps to prefer critical visual embedding over hallucinated words.

To address the problems with solely generalist backbone, a multi-agent system [10] can be a great solution. But, their architecture commonly depend on non-adaptive, static and rigid frameworks. Finally, to surmount problems of multi-agent pipelines, a group relative policy optimization (GRPO) framework [11], [12] can be enhanced with intra-group relative reward model, avoiding the necessity of a critic model. To optimize co-operation between the expert model and assistant model, curriculum guided reinforcement learning (C-RL) [13] with GRPO can be a viable solution. Gradually, coaching the model through the exploration and exploitation dilemma on difficult cases, which leads to hallucinations.

In this manuscript, we propose an integrated clinical agent that relies on LLM model orchestration with curriculum-based reinforcement learning (RL). Our key contributions are, (a) A multimodal alignment architecture with reasoning, lesion segmentation module and use a weighted DPO framework guided by diagnostic relevance ranking; (b) Implementing a clinical-aware preference optimization (CPO) framework with local lesion-noising to improve visual and textual alignment; and (c) A agent collaboration with GRPO based curriculum learning (C-GRPO) to surpass prominent generalist model, GPT-5 [14] in

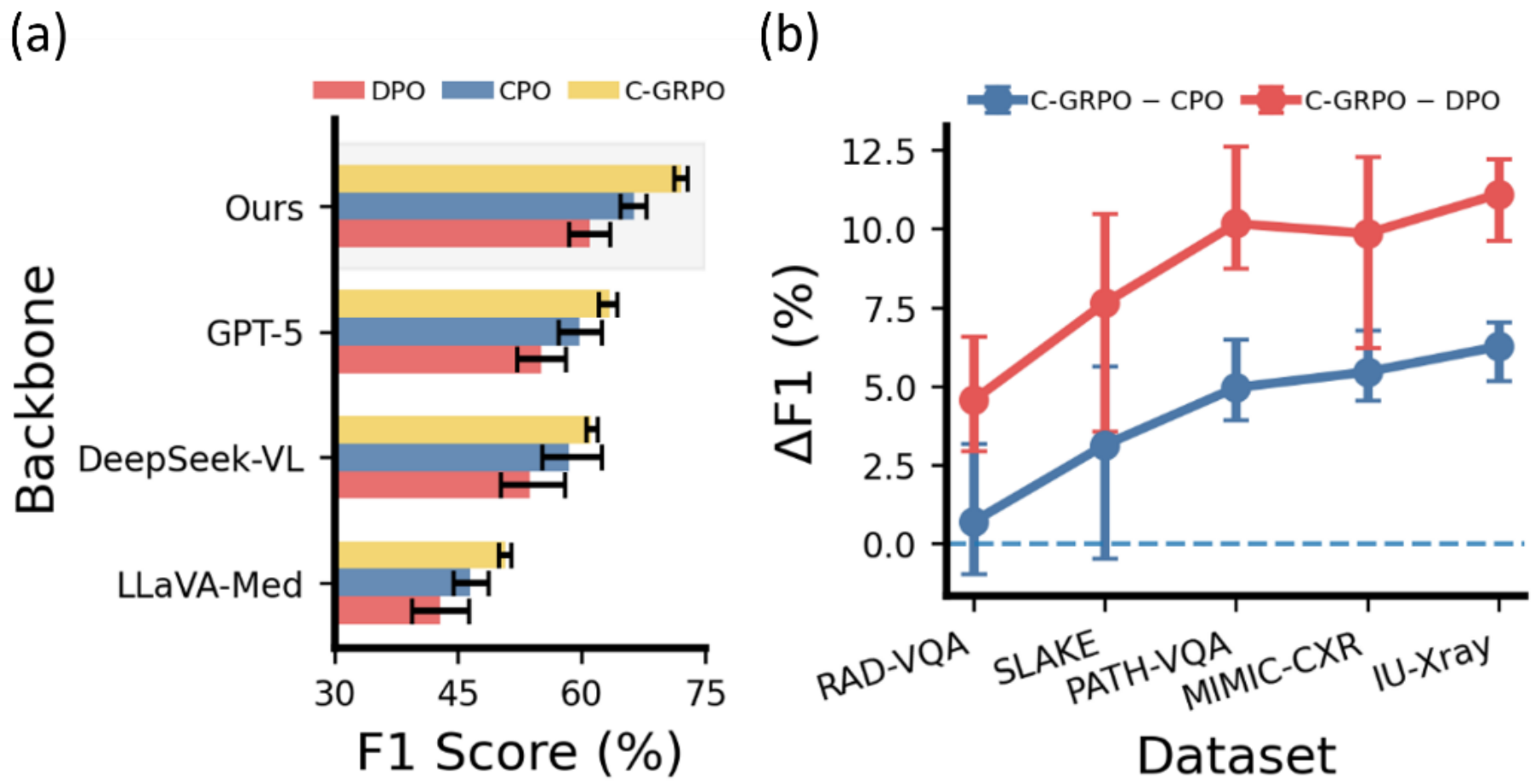

Fig. 2. The performance evaluation between DPO, CPO and C-GRPO- (a) architecture-specific; and (b) Dataset-specific performance efficiency gains of C-GRPO RL approach.

**Algorithm 1** C-GRPO RL Algorithm

**Require:** Dataset $\mathcal{D} = \{(x_{vis(i)}, x_{t(i)}, y_{*(i)})\}_{i=1}^{N}$, policy model $\pi_{\theta}{}_{EA}^{agg}$, old policy $\pi_{\theta_{\text{old}}}$, group size $N$, specialist responses $y_{r_d(i)}$

**Ensure:** $\pi_{\theta}{}_{EA}^{agg}$

1: Initialize $\mathcal{D}_{\text{easy}}, \mathcal{D}_{\text{medium}}, \mathcal{D}_{\text{hard}} \leftarrow \emptyset$
2: **for all** $(x_{vis}, x_t, y_*) \in \mathcal{D}$ **do**
3: $\quad s \leftarrow \text{Accuracy}(y_{r_d}, y_*)$
4: $\quad$ **if** $s = 1$ **then**
5: $\quad\quad \mathcal{D}_{\text{easy}} \leftarrow \mathcal{D}_{\text{easy}} \cup \{(x_{vis}, x_t, y_*)\}$
6: $\quad$ **else if** $0 < s < 1$ **then**
7: $\quad\quad \mathcal{D}_{\text{medium}} \leftarrow \mathcal{D}_{\text{medium}} \cup \{(x_{vis}, x_t, y_*)\}$
8: $\quad$ **else**
9: $\quad\quad \mathcal{D}_{\text{hard}} \leftarrow \mathcal{D}_{\text{hard}} \cup \{(x_{vis}, x_t, y_*)\}$
10: $\quad$ **end if**
11: **end for**
12: **for all** $(x_{vis}, x_t, y_*) \in \mathcal{D}$, $\mathcal{D} \in \{\mathcal{D}_{\text{easy}}, \mathcal{D}_{\text{medium}}, \mathcal{D}_{\text{hard}}\}$ **in batch do**
13: $\quad$ Sample $N$ rollouts $\{y_{\text{final}}^{(i)}\}_{i=1}^{N}$:
$\quad y_{\text{final}}^{(i)} \sim \pi_{\theta_{\text{old}}}(y \mid (x_{vis}, x_t), y_{r_d})$
14: $\quad$ **for** $i \leftarrow 1$ **to** $N$ **do**
15: $\quad\quad r_i \leftarrow r_{\text{format}}\left(y_{\text{final}}^{(i)}\right) + r_{\text{accuracy}}\left(y_{\text{final}}^{(i)}\right)$
16: $\quad$ **end for**
17: $\quad$ Compute groupwise advantage:
$\quad A_i \leftarrow \dfrac{r_i - \text{mean}(\{r_j\}_{j=1}^{N})}{\text{std}(\{r_j\}_{j=1}^{N})}$
18: $\quad$ Compute entropy regularization:
$\quad H_t \leftarrow -\sum_{j=1}^{T} p_{t,j} \log p_{t,j}$
19: $\quad$ Compute loss:
$\quad \mathcal{L}_{\text{C-GRPO}}(\theta) = E\left[\mathcal{L}_{\text{GRPO}}(\theta) + \gamma_s H_t(\pi_{\theta_{\text{agg}}}^{EA})\right]$
20: $\quad$ Update $\pi_\theta$
21: **end for**
22:

complex clinical and grounding across various metrics and benchmarks. The more details about agent backbone paradigm is depicted in Figure 1(b). The model acquires the ability to wisely combine contradictory recommendations and trust in intrinsic reasoning, while the expert entity collapses. This integrated architecture founds a new standard for automated intelligence systems.

## II. Prior Works

### A. Clinical AI Agent

Recently, AI agent has emerged with great prospects in clinical applications and reasoning. Some are unimodal, and some are multimodal in nature. AI agent expertise in medical setups [15], [16], [17], such as clinical, radiology, pathology, radiotherapy, dermatology, dental, genomics, EHR, and surgical have gained commendable recognition in the research arena. But most of the agents are either unimodal (text-based) or perform better in textual reasoning. Moreover, they are prone to multimodal misalignment, unstable and computation-intensive, limited generalization, a static sole backbone with no self-evolving, no adaptation, and no re-correction mechanism. As a result, they underperform by causing factual misalignment, false contextual reasoning, inefficient lesion grounding and segmentation, and hallucination [3].

### B. Reinforcement Learning Based Agent

On the other hand, reinforcement learning (RL) is used as a great tool in the agentic AI space, due to its self-adaptation, iterative, and self-evolving mechanisms. RL concepts [18] like, DPO, PPO, CPO, and GRPO, etc. are used in generalist tasks like, dialogue continuation, report generation, captioning, self-reasoning, spatial recognition, and vision cue generation related to medical fields. But, they often suffer inaccurate processes with human feedback alignment, instability, a generalist backbone, and in-adaptively orchestrate LLM tools, non-iterative and mismatched co-operation between expert tools [19] (refer to Figure 1(b)).

TABLE I The performance comparison of RL Approaches across 5 clinical benchmarks.

| Metrics and Checkpoints | | | | | | |
|---|---|---|---|---|---|---|
| **Backbone** | **RL Methods** | **RAD-VQA** | **SLAKE** | **PATH-VQA** | **MIMIC-CXR** | **IU-Xray** |
| **LLaVA-Med** | **DPO** | 47.54 | 41.65 | 40.29 | 47.37 | 37.42 |
| | **CPO** | 50.21 | 45.87 | 45.49 | 47.59 | 43.93 |
| | **C-GRPO** | **50.43** | **51.31** | **49.59** | **51.9** | **49.95** |
| **DeepSeek-VL** | **DPO** | 56.2 | 61.12 | 51.71 | 49.94 | 49.79 |
| | **CPO** | 61.19 | 65.12 | 55.73 | 55.56 | 55.52 |
| | **C-GRPO** | **60.77** | **62.65** | **60.45** | **60.89** | **60.91** |
| **GPT-5** | **DPO** | **60.26** | **57.1** | **52.36** | **50.51** | **55.18** |
| | **CPO** | **64.6** | **61.45** | **57.34** | **55.8** | **60.46** |
| | **C-GRPO** | **63.23** | **65.19** | **61.09** | **63.22** | **64.02** |
| **Ours** | **DPO** | 65.29 | 60.43 | 56.8 | 61.33 | 60.77 |
| | **CPO** | 68.72 | 65.9 | 63.43 | 67.79 | 66.61 |
| | **C-GRPO** | **73.08** | **71.71** | **70.68** | **72.55** | **72.68** |

## III. METHODS

In this part, we provide a concise overview of VLLMs, multimodal agent collaboration, DPO, CPO, and GRPO techniques with dynamic entropy regulation.

**VLLM:** Let multimodal clinical input $x_{in} = (x_{vis}, x_t)$, where, visual input is $x_{vis}$ and corresponding text input is $x_t$. They auto- iteratively guess the following token's distribution to generate clinical reasoning and grounding response $y_{out}$.

**Multi-Agent System:** To help with complicated agent pipelines, multi agent clinical systems harmonize various expert agents. Our scenario imitates a clinical setup for radiology diagnostics: Expert Agent (EA) → Clinical Specialist → Expert Model (EA). If an expert agent $a^i \in A^N$ observe an policy method $\pi_{\theta_i}(x_{in} | y_{out})$ for a multimodal vision cue $x_{in} = (x_{vis}, x_t)$, the multimodal agent systems would be, $A_{EA}^N = \{a_{EA}^1, a_{EA}^2, a_{EA}^3, ,,,,,,, a_{EA}^N\}$. Where, N = total number of agent and $\pi_{\theta_i}$ is the policy model. The pipeline progresses as follows: (a) Task captioning: $a_{EA}^{cap}$ agent selects the category of task based on the task caption and category of tasks via ranking scores, $r_d = arg\, max_k\, \pi_{\theta_{EA}^{cap}}(k \mid x_{in})$; and (b) Expert model response: Every individual captioning agent $a_{EA}^{r_d}$ after ranking most-k input sequence, and select the corresponding expert agent from the pool, thus produce responses, $y_{r_d} \sim \pi_{\theta_{EA}^{r_d}}(y_{out} \mid x_{vis})$; and (c) Aggregation agent: The output from expert agent, then fend into aggregator ($a_{EA}^{agg}$) to choose and finalize the appropriate output, $y_{final} \sim \pi_{\theta_{EA}^{agg}}(y_{out} \mid x_{in}, y_{r_d})$.

**Direct Preference Optimization (DPO):** Preference optimization has demonstrated to be very effective in fine-tuning language models (LMs) and leads to substantial coordination between target and model behavior. If a DPO [5] policy method $\pi_{\theta_{DPO}}$ produces a contingent distribution $\pi_{\theta_{DPO}}(y_{out} | x_{in})$ across a preference data $D = \{x_{in}, y_{pref}, y_{dlike}\}$, where $x_{in}$, $y_{pref}$, and $y_{dlike}$ represent the multimodal input, preferred response, and disliked output, respectively. The likelihood score of choosing $y_{pref}$ is designed as, $S\left(y_{pref} \mid y_{dlike}\right) = \sigma\left(r\left(x_{in}, y_{pref}\right) - r(x_{in}, y_{dlike}\right)$. Where, $\sigma(.)$ = the sigmoid non-linear function. The DPO optimization can be calculated through loss function over preference data $D$, $\mathcal{L}_{DPO}\left(\pi_{\theta_{DPO}}, \pi_{\theta_{pref}}\right) = -E_{(x_{in}, y_{pref}, y_{dlike}) \sim D}\left[\log\sigma\left(\alpha\log\frac{\pi_{\theta_{DPO}}(y_{pref} \mid x_{in})}{\pi_{\theta_{pref}}(y_{pref} \mid x_{in})}\right) - \alpha\log\frac{\pi_{\theta_{DPO}}(y_{dlike} \mid x_{in})}{\pi_{\theta_{pref}}(y_{dlike} \mid x_{in})}\right]$ Here, $\pi_{\theta_{pref}}$ = preference policy.

**Context-aware Preference Optimization (CPO):** During DPO, the likelihood score (S) is used as a penalty representing the influence of individual preference-pair data on the total fine-tuning process. To avoid underfitting caused by a small value from the DPO loss function, normalization is applied to achieve better context-aware preference. It maps every score within a fixed range with its variance and mean intact. The normalized clinical relevance score would be, $\bar{S} = \frac{S-\mu}{\sigma}$. Where, $(\mu, \sigma)$ are the mean and variance pair. Then the score is trimmed into $[\alpha, \beta]$. Here, $\alpha, \beta$ are chosen as lower and upper bound respectively. So the modified weighted loss function with diagnostic relevance can be computed as,

$$\mathcal{L}_{CPO}\left(\pi_{\theta_{CPO}}, \pi_{\theta_{pref}}\right) = -E_{(x_{in}, x_*, y_{pref}, y_{dlike}, \bar{S}) \sim D}\left[\bar{S}\log\sigma\left(\alpha\log\frac{\pi_{\theta_{DPO}}(y_{pref} \mid x_{in})}{\pi_{\theta_{pref}}(y_{pref} \mid x_{in})}\right) - \alpha\log\frac{\pi_{\theta_{DPO}}(y_{dlike} \mid x_*)}{\pi_{\theta_{pref}}(y_{dlike} \mid x_*)}\right]$$

**Group Relative Policy Optimization (GRPO):** The GRPO method [11] escapes fine-tuning a critic function by utilizing within-group relative rewards to enhance the policy function. For each multimodal inquiry $x_{in}$ , the pipeline extracts $N$ output responses $\{y_1, y_2, ,,,,,, y_N\}$, which are rated to obtain a list of rewards $\{r_1, r_2, ,,,,,, r_N\}$. It always chooses and contains normalized leverage points and refreshes the policy function with a trimmed objective function similar PPO [4]. The GRPO optimization would be, $\mathcal{L}_{GRPO}(\theta) = E_{(x_{in}, y_i)}\left[\frac{1}{N}\sum_i^N(\min(r_i A_i, clip(r_i, 1-\alpha, 1+\alpha) A_i) - \beta D_{KL}\left(\pi_{\theta_{GRPO}} \mid \pi_{\theta_{pref}}\right)\right]$. Here, each reward component would be, $r_i = \frac{\pi_{\theta_{GRPO}}(y_i \mid x_{in})}{\pi_{\theta_{old}}(y_i \mid x_{in})}$, $\pi_{\theta_{old}}$ = old GRPO policy function, and $\alpha, \beta$ are the agent's hyper-parameters, respectively. So, the final agent advantage component would be, $A_i = \frac{r_i - \mu(r_j \mid_{j=0}^N)}{\sigma(r_j \mid_{j=0}^N)}$. The GRPO enhance policy based learning through grouped relative policy rewards rather than a critic function.

**Curriculum Design (C-GRPO) and Dynamic Entropy Regulation:** We classify jobs based on the performance of an expert's assessment, $S = ACC\left(y_{r_d}, y_*\right)$. The whole dataset can be split into 3 complexity levels: totally accurate expert outputs (*S=1,* tagged as easy ($D_{Easy}$)), partially accurate results ($0 < S < 1$, marked as medium ($D_{Medium}$)), and entirely incorrect findings (*S = 0, designated as hard ($D_{Hard}$)*). So, the full dataset would be, $D = D_{Easy} \cup D_{Medium} \cup D_{Hard}$. Therefore, we create a triad-step curriculum based on a GRPO approach to enhance expert knowledge, like when to apply expert knowledge precisely and when to depend on their individual perception to resolve multimodal cues.

However, we apply GRPO [11] as our base RL curriculum approach. For each multimodal query *x*, aggregation policy $\pi_{\theta_{EA}^{agg}}$ producing a collection of *N* responses $\{y_{final}^{i}\}_{i=1}^{N}$. Each reward component $r_i$ for each output is calculated by a reward template $r_{format} \in \{0, 0.5\}$ and reward accuracy $r_{accuracy} \in \{0, 1\}$, and computes the conditional advantage component $A_i$ for policy model update. To attain dynamic entropy control, we present an entropy regularization into our GRPO optimization function,

$$\mathcal{L}_{C-GRPO}(\theta) = E\left[\mathcal{L}_{GRPO}(\theta) + \gamma_s H_t\left(\pi_{\theta_{EA}^{agg}}\right)\right] \text{--(1)}$$

Where, $H_t = -\sum_{j=1}^{T} p_{t,j} \cdot log p_{t,j}$.

And, $p_t = Softmax\left(\frac{Z_t}{\tau}\right)$, T = vocabulary token size, softmax logits $Z_t \in R^T$, $\tau$ = temperature, $\gamma_s$ = entropy penalty co-efficient is adaptively established based on curriculum level *S* of current cues. To mention, $\gamma_{easy} \sim 0$, $\gamma_{medium} > 0$, and $\gamma_{hard} \sim 1$ (>> $\gamma_{medium}$) respectively for easy ($S = 1$), medium ($0<s<1$), and hard ($S = 0$) cues. In case of hard cues, a very strong positive penalty is implied vigorously to stimulate exploration, forcing the agent to deviate from deceptive expert suggestions.

**Initial Captioning Expert:** Contemporary task-specific agent approaches like Agent Hospital [20], depend on preset or rule-based allocation plans and do not directly improve the captioning mechanism. Therefore, expert-based routing is fixed and misses flexible optimization over the clinical data. To handle those drawbacks, we formulate our initial task captioning module as a tunable policy $\pi_{\theta_{EA}^{cap}}$ that allocates every instance to one of *K* domain experts, where *K* = 7 for our BioMed-Agent-RL. Supervising is obtained from method-dependent ground-truth offered by dataset (e.g., radiology image → radiologist LLM backbone), allowing methodical positioning between input and medical expertise. Then, the chosen caption can be denoted based on ranking, tasks via ranking scores, $r_d = arg\,max_k\, \pi_{\theta_{EA}^{cap}}(k \mid x_{in})$, where x = multimodal clinical cue. We improve this strategy utilizing GRPO with a combined reward model comprising of template restrictions for report generation $\left(r_{format\,/\,caption} \in \{0.0, 0.50\}\right)$ and reward accuracy$\left(r_{accuracy/\,F1} \in \{0.0, 1.0\}\right)$. This kind of reward model promotes assigning of both precise specialty and systematic semantic reasoning results.

**Role Specialist:** After choosing appropriate captions, we employ various robust pipelines as field specialists $a_{EA}^{r_d}$ to facilitate moderately precise evaluations. This assists following reference by a human expert. In our framework, we utilize outputs from *K* specialists as pointers for every input sample. We collect the human expert opinions to autonomously incorporate in the data and backbone. So expert output would be, $y_{r_d} \sim \pi_{\theta_{EA}^{r_d}}(y_{out} \mid x_{in})$. This guarantees framework efficacy and bypasses dominant polling, which could dominate subordinate views and assign the task for final call by a physician.

**Aggregation Agent:** After expert-tier reasoning is produced, the ultimate investigative judgement could be taken by a medical expert who combines diverse specialist views to the final output. In this integration stage, both human and agent expert outputs may vary in terms of reliability or, disagree with one another. However, majority voting may result in depriving minority explanations may cause propagation of structured errors. Therefore, we are proposing adaptive aggregation policy with entropy regulation, [21] policy-based to reveals attending physician in complex clinical cases, thus empowering systemic adaptation. This policy C-GRPO entropy controls the exploitation and exploration balance. To add more, low entropy fosters agreement-based exploitation while expert consensus is credible. However, high entropy promotes exploratory thinking when unpredictability exists. By incorporating entropy regulation to RL pipeline, C-GRPO reduces error buildup, imbalance between LLM experts, and improves consistency in final outputs.

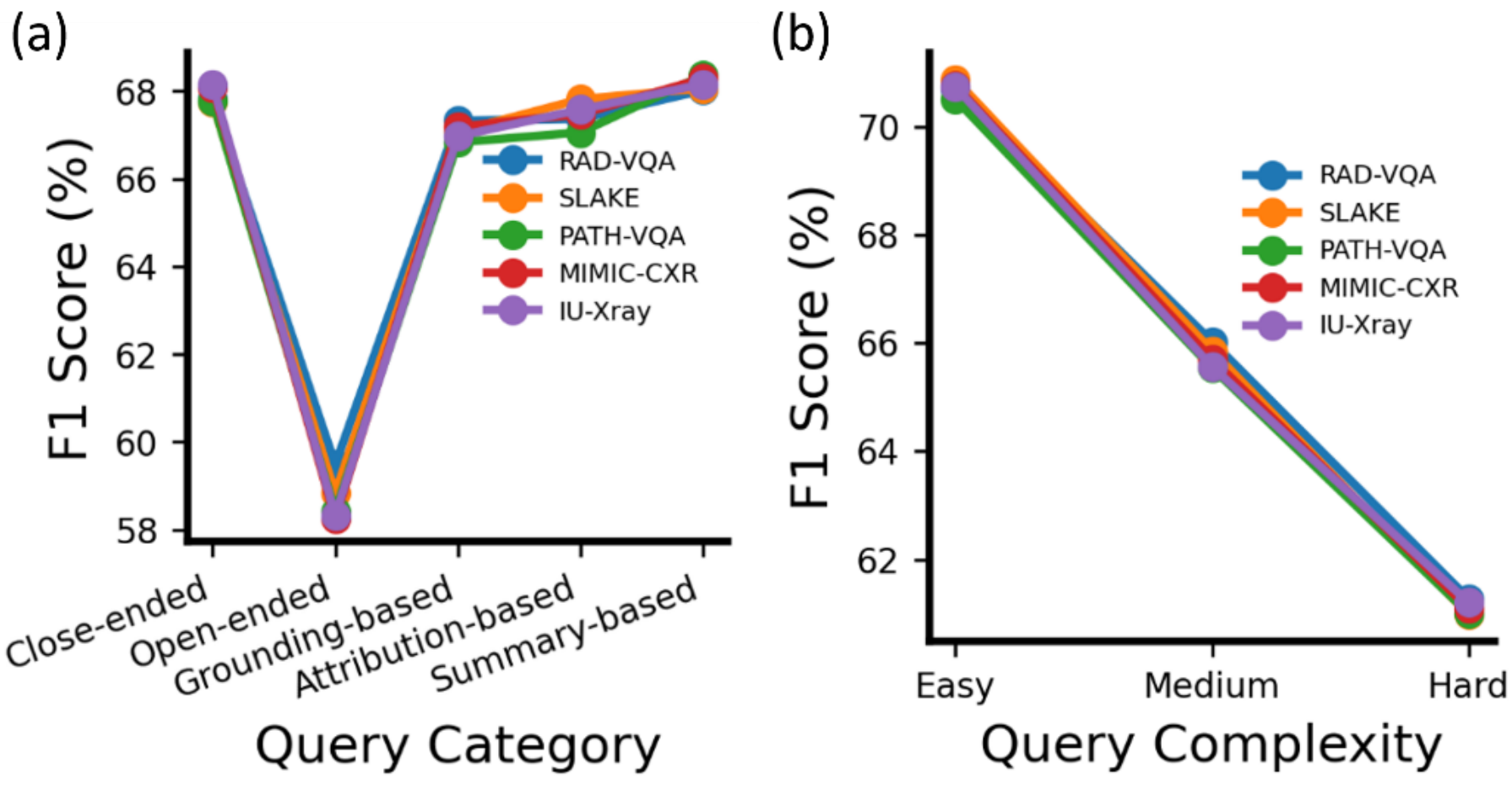

Fig. 3. The query category and complexity level analysis for our C-GRPO based BioMed-Agent-RL (1200 instances per category, and 2000 data points per complexity).

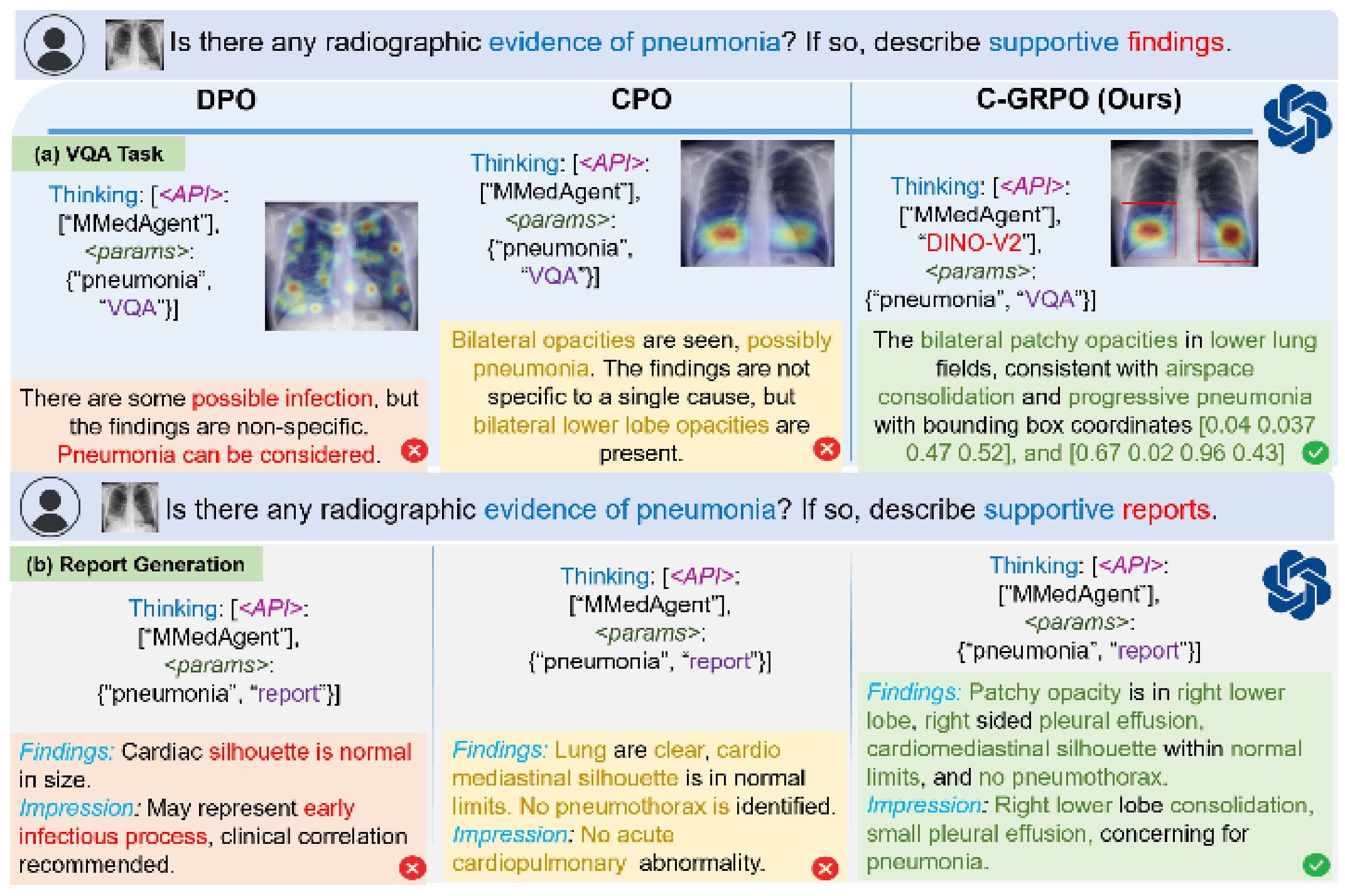

Fig. 4. The BioMed-Agent-RL qualitative result comparison for different RL Algorithms- (a) VQA task, and (b) Report generation task.

TABLE II THE QUERY COMPLEXITY ANALYSIS FOR C-GRPO BIOMED-AGENT-RL (6000 INSTANCES).

| Reasoning difficulty in various dimensions | | | | | | |
|---|---|---|---|---|---|---|
| Query Categories | Query Complexity | RAD-VQA | SLAKE | PATH-VQA | MIMIC-CXR | IU-Xray |
| Close-ended | Easy | 73.08 | 72.98 | 72.45 | 73.01 | 72.88 |
| | Medium | 68.83 | 68.55 | 68.69 | 68.59 | 68.67 |
| | Hard | 61.53 | 61.71 | 62.18 | 62.66 | 62.93 |
| Open-ended | Easy | 65.21 | 65.48 | 64.64 | 64.7 | 65.35 |
| | Medium | 59.67 | 58.11 | 57.83 | 58.03 | 57.97 |
| | Hard | 53.56 | 52.89 | 52.78 | 51.99 | 51.64 |
| Grounding-based | Easy | 70.94 | 71.79 | 71.86 | 71.91 | 71.56 |
| | Medium | 66.82 | 65.95 | 65.53 | 65.83 | 65.62 |
| | Hard | 64.22 | 63.49 | 63.11 | 63.82 | 63.74 |
| Attribution-based | Easy | 71.31 | 71.73 | 70.99 | 71.39 | 71.45 |
| | Medium | 67.51 | 67.9 | 66.79 | 67.12 | 67.26 |
| | Hard | 63.29 | 63.82 | 63.4 | 63.92 | 64.03 |
| Summary-based | Easy | 72.97 | 72.41 | 72.65 | 72.83 | 72.48 |
| | Medium | 67.32 | 68.79 | 68.89 | 68.95 | 68.33 |
| | Hard | 63.78 | 62.99 | 63.55 | 63.09 | 63.68 |

## IV. EXPERIMENTS

### A. Datasets and Baselines

We measure BioMed-Agent-RL over 2 agent

architectural setups: (a) Single Backbone, and (b) Domain-specific list of VLLMs experts. Particularly, we incorporate the performance evaluation of our proposed agent with LLaVA-Med [22], DeepSeek-VL [23], and GPT-5 [13]. We train and fine-tune our agent on 5 medical datasets upon 2 categories: (a) VQA domain- RAD-VQA [24], SLAKE [25]; and (b) Report generation- Path-VQA [26], MIMIC-CXR [27], and IU-Xray [28]. The evaluation query types are a mixture of multimodal closed-ended (e.g., yes/ no), open-ended (e.g., abnormality type), grounding-based (e.g., counting), attribution-based (e.g., shape, size, and density), Q&A, and summary generation tasks. We rigorously follow the standard for splitting the datasets into 40%: 30%: 30% for train, validation, and test sets, respectively. To promote a curriculum learning approach, we additionally grouped train and validation sets into prior categories- VQA domain (dialogue) and Report generation (explanation). And, each sub-categories further sampled into 3 complexity levels, easy, medium, and hard, based on the subject-expert's feedback. And, for DPO and CPO optimization, each data point contains preferred and non-preferred data fields. Overall, our experiment cases are combinations of 3 RL methods, 4 model backbones, 2 benchmark categories (5 datasets), 5 query categories, and 3 query complexity levels. In total, 3 x 4 x 5 x 5 x 3 = 900 experimental cases are examined retrospectively on the test data to illustrate the agent's robustness against specialist ground truth and confirm no potential data leak.

## B. Implementations

We employ MMedAgent [10] as a backbone and develop the prompt blueprint to clearly designate mandatory inference structure based on VQA and report generation query category. It involves using <*thinking*> and <*gpt_answer*> tag to individually enclose reasoning and lesion grounding process, and inference answer. More details about the workflow is depicted in Figure 1 (a). A curriculum based GRPO algorithm (C-GRPO) is implemented to make the agentic system more robust, accurate grounding, and less hallucinated results (refer to Algorithm 1). The agent's hyperparameters are- batch size- 256, rollouts- 8, temperature- 9.0, learning rate- $1x10^{-5}$, KL curriculum coefficients are- $3x10^{-3}$, $5x10^{-5}$, $7x10^{-7}$, # of specialists- 3, and dynamic entropy coefficients- 0.03, 0.003, 0.0003 respectively for query complexity (hard (s = 0), medium ($0<s<1$), and easy (s=1) multimodal cues). The baseline training and tuning are done on Nvidia 8xA5000, 192 GB VRAM. The whole process took 10~12 days straight to get the final agent checkpoints and inference results.

## C. Ablation Studies

In this part, we assess the overall performance of BioMed-Agent-RL, striving to solve subsequent challenges: (a) Does C-GRPO enhance comprehensive multimodal reasoning and grounding?, (b) Are the performance improvements robust across all datasets?, (c) Is the performance enhancement model architecture independent?, (d) does C-GRPO helped in all query categories and complexity levels?.

From Figure 2(a), the metrics of average F1 (%) across 3 RL algorithm at 4 diverse backbone, and combined over 5 clinical benchmarks. Throughout all architectures, C-GRPO uniformly attains better efficiency contrasted to CPO and DPO algorithms. The mean F1 (%) ranges from 49.95% for LLaVA-Med to 73.08% for C-GRPO based MMedAgent checkpoints (refer to yellow bar). The uncertainty bars (error bar window) indicate 95% intervals of confidence is figured across 5 benchmarks, showing steady performance increases. These performance outcomes denotes that C-GRPO offers framework-neutral improvements in mixed modality grounding and reasoning. On the other hand, Figure 2(b) exhibits dataset-specific efficiency gains of C-GRPO proportional to CPO, and DPO RL approaches. Involving all 5 biomedical benchmarks, C-GRPO generates steady affirmative improvements with average gains varying from 1% to 10.5% of $\Delta F1$ (%). During each iteration of RAD-VQA benchmarks, C-GRPO running performance improved from 2.9% to ~4% (blue error bars) while fine-tuning for the IU-Xray dataset. Similarly, while 3rd, 4th and 5th benchmark, the agent is consistently learning and improving grain F1(%) of ~5%, 4.8% and ~6.5% respectively. It can be noticed that the gain from DPO to C-GRPO (C-GRPO-DPO, refer to red curve) shows much higher than the CPO to C-GRPO (C-GRPO – CPO) curves, which denotes the actual precedence of performance in descending order, C-GRPO > CPO > DPO. More details about performance are shown in Table 1, while the best performance (F1 (%)) for each benchmark and backbone is marked with green text. It can be claimed that with the consistent fine-tuning of each benchmark, the model's reasoning, grounding, and generalization improved with steady upward non-linear trends.

Figure 3 examines our C-GRPO based BioMed-Agent-RL output across 5 query categories and 3 query complexities. As displayed in Figure 3(a), efficiency persists comparatively firm across query categories- close-ended, open-ended, grounding-based, attribution-based, and summarizer. We consider 6000 data points in total for this study and 1200 multimodal data instances for each category. To begin with, open-ended queries are normally much more complex and dynamic in clinical applications; therefore show a little downward slope (dipped) to 59% (F1) accuracy. Otherwise, all other categories remained relatively flat (between ~68% and 73.3%). This suggests that the suggested enhancement strategy scales reliably across diverse semantic multimodal cues rather than over-learned specific category. In the contrary, Figure 3(b), shows query complexity analysis over our C-GRPO based BioMed-Agent-RL. Here, we are considering 2000 clinical instances per complexities. It can be noted that a methodical decline in efficiency as clinical grounding and reasoning difficulty rises from easy to hard level cues. The accuracy of easy cues (73.08%) moves downward to hard cues (~58%), mirroring the drop across all benchmarks. Finally, the comparative efficiency study of query complexities and categories for our agent helps to establish a robust, reliable, explainable, and trustworthy clinical intelligence systems. More details about the performances on query level is depicted in Table 2.

Figure 4 displays the qualitative result comparison among various checkpoints with C-GRPO approach. The goal is to analyze the quality of inference from 2 group of tasks, such as (a) VQA task, (b) report generation. VQA task refers to regular days to day clinical tasks though user interaction and regular dialogues by utilizing execution parameters <params> as "VQA" to the backbone. On the other hand, report generation refers to generated inference through a specific format ("Findings:" and "Impressions :"). It also passes parameter (<params> as "report") to the backbone. When asking a cues like, "any evidence of pneumonia?" The agent generates VQA task, DPO and CPO shows surface level output for clinical x-ray cue with acute diseases (e.g, "pneumonia") such as, "possible infection", "early infection,", and "Bilateral opacities". Whereas, our BioMed-Agent-RL replies with sematic clinical keywords like, "patchy opacity", "right lower lobe", "pleural effusion", and "acute pneumonia", grounding and reasoning with rectangular bounding boxes with format [x1, y1, x2, y2]. Based on captioning LLM, the model's prompt builder refers to the backbone with tag <API> which containing model name, for instance "MMedAgent", "Dinov2". Then the referred model will be invoked according to the referred tags. Collectively, these findings suggest that BioMed-Agent-RL efficiency enhancements are not only consistently distributed, but also specifically noticeable in clinical challenging semantic reasoning and grounding scenarios.

## V. Future Direction

Even though, the BioMed-Agent-RL outperforms in critical clinical reasoning, grounding and segmentation task, it is not without flaws. First, It oftentimes prone to show low accuracy for long longitudinal multimodal cues. Therefore, a persistent memory augmentation with adaptive retrieval is necessary to deal with episodic cues. We have already planning to implement a persistent memory module for our agentic system. Secondly, when dealing with curriculum learning in GRPO, we have a plan to implement an uncertainty aware reward modelling and policy function to improve reliability of the agent.

## VI. Conclusion

In this manuscript, we propose a curriculum learning on GRPO reinforcement learning based clinical agent for complex biomedical VQA and report generation tasks. The efficient reasoning and grounding approach is proposed across 5 semantic query categories, several biomedical benchmarks, and 3 complexity levels. Our C-GRPO approach reliably surpass DPO and CPO optimization approach up to maximum accuracy of 73.03% (with maximum gain of 10.5% compared to its closest RL approach (DPO)). The backbone-level studies confirms that the performance gains are architecture and algorithm agnostic. Furthermore, dataset-specific further validated that enhancements were steadily positive across diverse clinical tasks, indicating robust data generalization. Query semantic categories and complexity-level evaluation reveals that the agent generalize well for any uncommon and super complex clinical cues. The direction indicates that C-GRPO improves reasoning consistency under multi-stage and structural reasoning conditions, therefore reducing error distribution through reward modelling in clinical AI systems.